\documentclass[letterpaper, 10 pt, conference]{ieeeconf}  % Comment this line out if you need a4paper

\IEEEoverridecommandlockouts                              % This command is only needed if 
\usepackage{amsmath} % assumes amsmath package installed
\usepackage{amssymb}  % assumes amsmath package installed
\usepackage{graphicx}
\usepackage{multirow}
\usepackage{booktabs}
\usepackage{tabularx}

\makeatletter
\let\NAT@parse\undefined
\makeatother
\usepackage[colorlinks]{hyperref}
\hypersetup{colorlinks=true,linkcolor=red,citecolor=blue,urlcolor=magenta}

\usepackage[table,dvipsnames]{xcolor} % for rowcolor

\definecolor{rblue}{rgb}{0,0.5,1}

\title{\LARGE \bf
Learning Fine-Grained Correspondence with Cross-Perspective Perception for Open-Vocabulary 6D Object Pose Estimation}

\author{Yu Qin$^{1}$, Shimeng Fan$^{3}$, Fan Yang$^{1}$, Zixuan Xue$^{1}$, Zijie Mai$^{1}$, Wenrui Chen$^{1}$,\\Kailun Yang$^{1,2,*}$, and Zhiyong Li$^{1,*}$%
\thanks{This work was partially supported by the National Natural Science Foundation of China under Grant No. U23A20341, 62273137, 62473139, and No. U21A20518, the National Key R\&D Program of China under Grant 2022YFB4701400/2022YFB4701404, the Hunan Provincial Research and Development Project under Grant 2025QK3019, the Hunan Science Fund for Distinguished Young Scholars under Grant 2024JJ2027, the Special Funds for Construction of Innovative Provinces in Hunan Province under Grant 2025QK1005, and the State Key Laboratory of Autonomous Intelligent Unmanned Systems (the opening project number ZZKF2025-2-10).
\textit{(Corresponding authors: Kailun Yang and Zhiyong Li.)}
}
\thanks{$^{1}$The authors are with the School of Artificial Intelligence and Robotics and the National Engineering Research Center of Robot Visual Perception and Control Technology, Hunan University, Changsha 410082, China. (E-mail: kailun.yang@hnu.edu.cn, zhiyong.li@hnu.edu.cn.)}%
\thanks{$^{2}$The author is also with the State Key Laboratory of Autonomous Intelligent Unmanned Systems, Tongji University, Shanghai 201804 China.}%
\thanks{$^{3}$The author is with the School of Computer Science and Engineering, Hunan University of Science and Technology, Xiangtan 411201, China.}%
}

\begin{document}

\maketitle
\thispagestyle{empty}
\pagestyle{empty}

%%%%%%%%%%%%%%%%%%%%%%%%%%%%%%%%%%%%%%%%%%%%%%%%%%%%%%%%%%%%%%%%%%%%%%%%%%%%%%%%
\begin{abstract}
Open-vocabulary 6D object pose estimation empowers robots to manipulate arbitrary unseen objects guided solely by natural language. However, a critical limitation of existing approaches is their reliance on unconstrained global matching strategies. In open-world scenarios, trying to match anchor features against the entire query image space introduces excessive ambiguity, as target features are easily confused with background distractors. To resolve this, we propose Fine-grained Correspondence Pose Estimation (FiCoP), a framework that transitions from noise-prone global matching to spatially-constrained patch-level correspondence. To systematically eliminate background interference, FiCoP first employs an object-centric disentanglement step to isolate the target from macro-level environmental noise. Building upon this localized region, our core methodological innovations are twofold. Firstly, a Cross-Perspective Global Perception (CPGP) module is proposed to fuse dual-view features, establishing structural consensus through explicit context reasoning and text-guided semantic injection. Secondly, we design a Patch Correlation Predictor (PCP) that leverages a patch-to-patch correlation matrix as a structural prior. This generates a precise block-wise association map, acting as a spatial filter to enforce fine-grained, noise-resilient matching. Experiments on the REAL275 and Toyota-Light datasets demonstrate that FiCoP improves Average Recall by 8.0\% and 6.1\%, respectively, compared to the state-of-the-art method, highlighting its capability to deliver robust and generalized perception for robotic agents operating in complex, unconstrained open-world environments. The source code will be made publicly available at \url{https://github.com/zjjqinyu/FiCoP}.
\end{abstract}

%%%%%%%%%%%%%%%%%%%%%%%%%%%%%%%%%%%%%%%%%%%%%%%%%%%%%%%%%%%%%%%%%%%%%%%%%%%%%%%%

\section{Introduction}
In the pursuit of general-purpose robotics, 6D object pose estimation is the prerequisite for enabling agents to interact with the physical world. While traditional methods rely on pre-scanned CAD models~\cite{pos3r, tta_cope, coop}, the field is shifting towards open-vocabulary paradigms enabled by Vision-Language Models (VLMs)~\cite{clip, siglip}, allowing robots to perceive novel objects via textual descriptions. 
However, achieving robust pose estimation in unconstrained environments remains a formidable challenge, particularly when there are large viewpoint differences between the reference (anchor) and the current observation (query).

\begin{figure}[!t]
\centering
\includegraphics[width=0.9\linewidth]{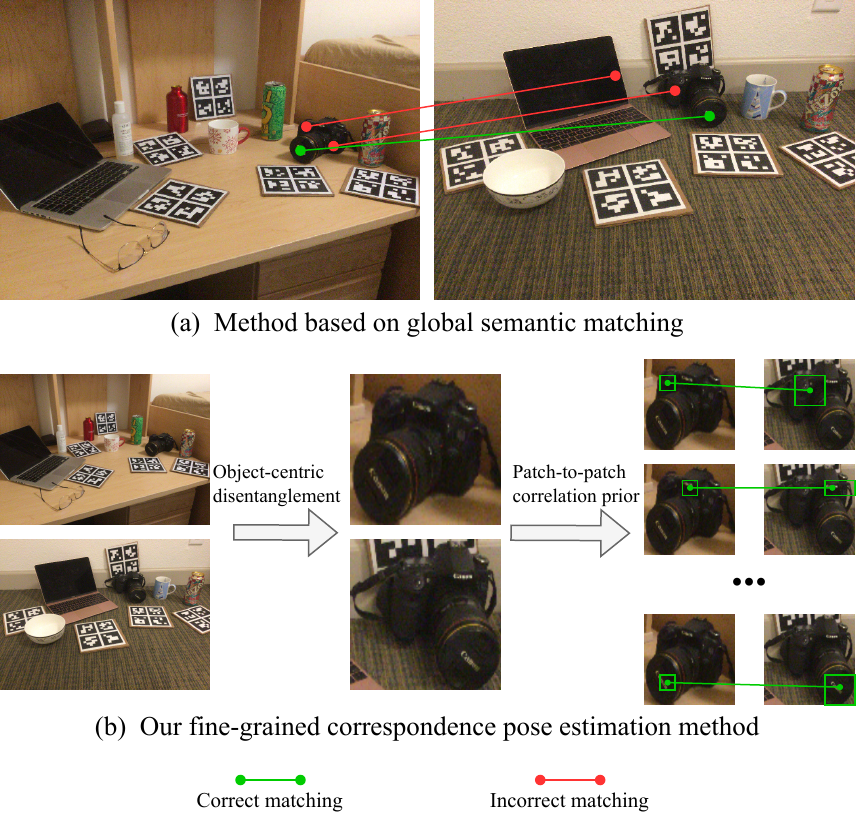}
\vskip-2ex
\caption{Comparison between our method and previous methods. 
Previous methods~\cite{oryon, horyon} based on global matching are prone to incorrect matching. 
Our method gradually refines the matching area through object-centric disentanglement and patch-to-patch correlation priors, aiming to promote more accurate matching.
}
\label{fig_intro}
\vskip-3ex
\end{figure}

Existing open-vocabulary methods~\cite{oryon, horyon} typically approach this problem through global semantic matching. 
They extract features from the anchor and attempt to match them with the entire query image. 
From the perspective of robotic manipulation, such an unconstrained search is inherently challenging in cluttered scenes. 
When the viewpoint changes drastically, the object's appearance deforms, and simple global similarity scores fail to distinguish the object parts from the changing appearance and similar-looking background noise, as shown in Fig.~\ref{fig_intro}~(a). 
This leads to a high error rate in pose prediction, as pixels in the anchor are often incorrectly matched to clutter in the query view. 
Regarding the inherent limitations of multimodal features, VLM representations suffer from a granularity mismatch: they are excellent at capturing high-level semantics but lack the spatial discrimination during global matching. Without a mechanism to constrain the search space, the distractors overwhelm the geometric information of the object.

To overcome these issues, we draw inspiration from the cognitive mechanisms of humans. 
When humans re-identify an object from a new angle, we do not scan every detail against the entire scene.
Instead, we perform a hierarchical process~\cite{human_vision}: we first locate the position of the object, then focus on relevant local regions that share structural similarity, and finally establish precise correspondences within those focused areas. 
This coarse-to-fine attention mechanism naturally filters out irrelevant visual information. 
Guided by this insight, this letter proposes FiCoP for \textbf{Fine-Grained} \textbf{Correspondence} \textbf{Pose} Estimation. 
Unlike previous methods~\cite{oryon, horyon} that perform coarse-grained global matching, FiCoP introduces a fine-grained correspondence learning mechanism, as shown in Fig.~\ref{fig_intro}~(b). 
The core concept is to utilize a patch-to-patch correlation matrix as a strong spatial prior. Instead of allowing an anchor pixel to match with any query pixel, our model first locates the object and then predicts which local patches in the query are structurally correlated with the anchor. This narrows the matching scope significantly. 
If an anchor patch is predicted to correspond only to the query patch, the subsequent pixel-level matching is confined within this region, automatically suppressing interference from the rest of the image.

To comprehensively evaluate the effectiveness of our approach, we have conducted extensive evaluations on the REAL275~\cite{nocs} and Toyota-Light~\cite{bop} benchmarks. 
The results demonstrate that FiCoP establishes a new state-of-the-art in open-vocabulary pose estimation, outperforming existing methods~\cite{oryon, horyon, posediff, realpose++, objectmatch, sift, latentfusion} by a substantial margin.
Specifically, on the REAL275 and Toyota-Light datasets, our method achieves an average recall improvement of 8.0\% and 6.1\%, respectively, compared to the state-of-the-art Horyon method that relies on global matching.
Crucially, qualitative analysis confirms that our fine-grained strategy successfully recovers accurate pose parameters even in scenarios with drastic viewpoint alterations where traditional global features typically collapse. 
These findings support our hypothesis that imposing spatial constraints via patch-level priors can effectively mitigate background ambiguity and enhance robust perception in unconstrained environments.

The main contributions of this letter can be summarized as follows:
\begin{itemize}
\item We propose FiCoP, a novel open-vocabulary 6D pose estimation framework that shifts the paradigm from noise-prone global semantic matching to a coarse-to-fine correspondence mechanism. This global-to-local approach effectively mitigates background ambiguity, facilitating more reliable pose estimation in unconstrained environments.

\item We design a Patch Correlation Predictor (PCP) as the engine of our fine-grained strategy. 
Specifically addressing the unconstrained global matching ambiguity in cluttered robotic scenes, the PCP acts as a 3D-aware spatial filter.
It explicitly computes the patch-wise similarity map to generate a spatially constrained search region, ensuring that final pixel correspondences are both semantically correct and geometrically precise.

\item We propose a Cross-Perspective Global Perception (CPGP) module, a transformer-based architecture fundamentally designed to bridge the geometric perspective gap between anchor and query views. By facilitating explicit cross-view feature interaction, CPGP establishes structural consensus under large 3D deformations. Additionally, CPGP auxiliarily injects textual semantics to suppress background clutter and focus structural alignment on the target object.
\end{itemize}

\section{Related Work}
\subsection{6D Pose Estimation in Classic Settings}
The classic 6D pose estimation methods are generally categorized into instance-level, category-level, and unseen objects.
Instance-level methods, such as DenseFusion~\cite{densefusion}, RCVPose~\cite{rcvpose}, SCFlow~\cite{scflow}, and Uni6D~\cite{uni6d}, achieve high precision by establishing geometric correspondences or direct regression but are limited to determining the pose of specific instances seen during training. 
To generalize across instances, category-level methods~\cite{nocs, sgpa, dualposenet} learn shared canonical representations to estimate poses for objects within known categories. However, they rely heavily on category-specific shape priors. 
For unseen objects, methods like Megapose~\cite{megapose} and Gen6D~\cite{gen6d} employ render-and-compare optimization or reference image matching. While generalizing better, they still necessitate CAD models or multi-view references, restricting their utility in unstructured environments. 

Unlike these traditional approaches that rely heavily on explicit CAD models or category-specific shape priors, our method adopts a model-free paradigm driven solely by textual descriptions, enabling true zero-shot generalization in the open world.

\begin{figure*}[t]
\centering
\includegraphics[width=0.9\linewidth]{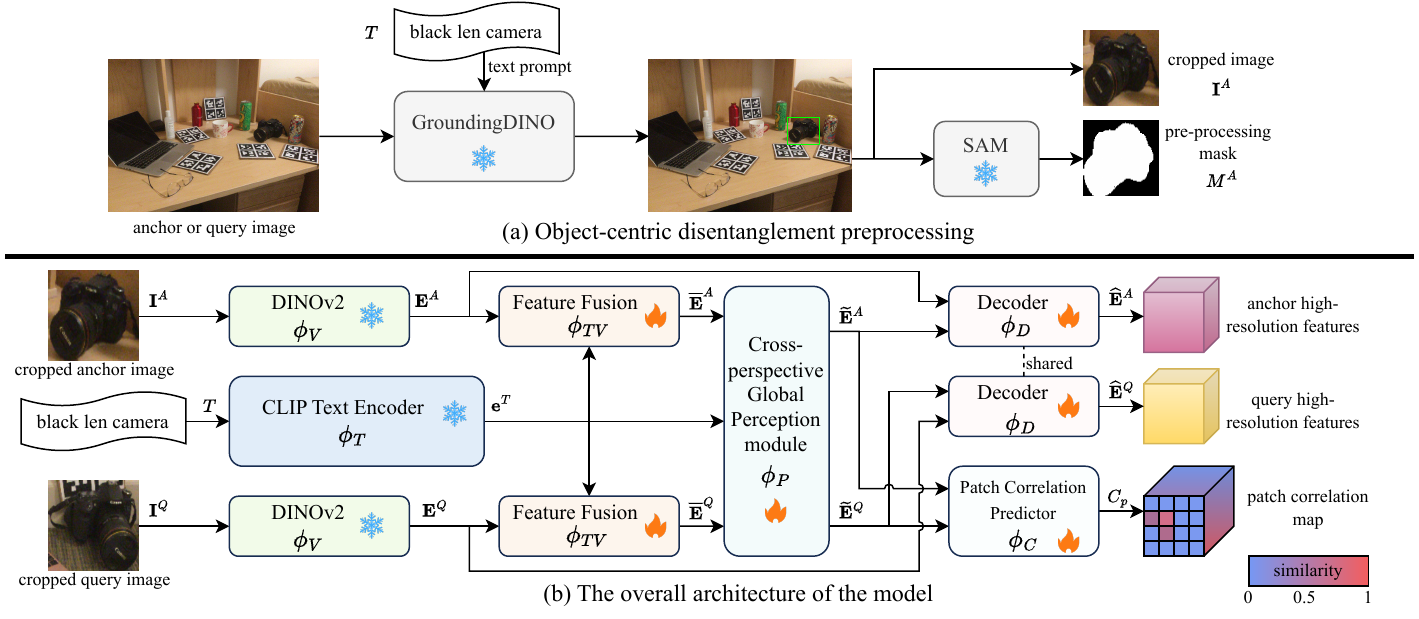}
\vskip-2ex
\caption{The framework of our proposed Fine-grained Correspondence Pose Estimation (FiCoP) model.
It consists of two stages: (a) a preprocessing pipeline that utilizes an open-vocabulary object detection model and a SAM model to generate cropped object images and masks; (b) a model forwarding process that takes the preprocessing results as input to generate high-resolution features for anchor and query, as well as patch correlation maps.}
\label{fig_overall}
\vskip-2ex
\end{figure*}

\begin{figure}[t]
\centering
\includegraphics[width=0.9\linewidth]{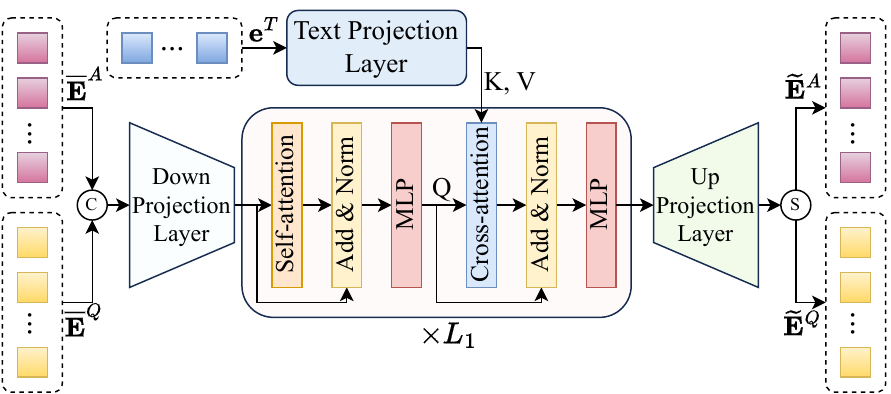}
\vskip-2ex
\caption{Structure of the Cross-Perspective Global Perception (CPGP) module. This module facilitates information interaction between the anchor and query perspectives.}
\label{fig_cpgp}
\vskip-2ex
\end{figure}

\subsection{Open-Vocabulary 6D Pose Estimation}
The advent of large-scale Visual Language Models (VLMs) like CLIP~\cite{clip} has shifted computer vision from closed-set recognition to open-vocabulary learning. 
In 6D pose estimation, category-level methods such as OV9D~\cite{ov9d} and LightPose~\cite{lightpose} utilize cross-modal knowledge to generalize across textual descriptions, yet they remain confined to known categories. 
The pioneering open-vocabulary paradigm for unseen objects, introduced by Oryon~\cite{oryon}, overcomes this by specifying objects solely through a textual prompt, using a CLIP-based fusion module to integrate semantic cues with local geometry for cross-scene segmentation and matching without any prior object model. 
The subsequent model Horyon~\cite{horyon} significantly advances this paradigm by addressing the critical limitations of low-resolution fusion features and background clutter. 
Due to the challenging nature of this paradigm, estimated pose accuracy is typically not very precise, especially when dealing with significant view differences between anchors and queries.
Additionally, relative pose estimation methods~\cite{posediff, realpose++, latentfusion}, and point cloud registration methods~\cite{objectmatch, sift, latentfusion} have been adapted for open-vocabulary pose estimation by incorporating external open-vocabulary object detectors or masks. 

However, these methods typically rely on unconstrained global semantic matching, which often degenerates into ambiguity when facing substantial viewpoint discrepancies or background clutter. 
In contrast, our work advances this paradigm by introducing a fine-grained patch-constrained mechanism and explicit cross-perspective interaction, effectively recovering the precise geometric structural consistency that is often lost in coarse global features.

\section{Methodology}
\label{sec_method}

\subsection{Problem Formulation}
We describe the open-vocabulary 6D object pose estimation problem: given an image pair consisting of a reference anchor $\mathbf{I}^A$ and a current observation query $\mathbf{I}^Q$, along with a natural language description $T$ of the target object, our goal is to predict the rigid transformation $\mathbf{T}_{A\to Q} \in SE(3)$. This transformation aligns the object's coordinate system from the anchor view to the query view. 
Unlike the traditional paradigm requiring CAD models~\cite{rcvpose, scflow, megapose, gen6d}, we rely solely on $T$ to identify the target, necessitating a robust mechanism to filter environmental distractors and establish precise correspondences across large viewpoint changes.

\subsection{Object-Centric Disentanglement Preprocessing}
To address the global matching ambiguity highlighted in the introduction, we first implement an Object-Centric Disentanglement Preprocessing. 
As shown in Fig.~\ref{fig_overall}~(a), this stage first employs GroundingDINO~\cite{groudingdino} to localize the target object in both anchor and query images based on the provided text prompt $T$. 
This module outputs bounding boxes that tightly enclose the object of interest. Subsequently, the Segment Anything Model (SAM)~\cite{sam} is applied to generate preliminary segmentation masks for the localized objects. 
This preprocessing step effectively isolates the object from background clutter while outputting high-quality object masks, which is critical for subsequent accurate pose estimation.

\subsection{Feature Extraction and Fusion}
As illustrated in Fig.~\ref{fig_overall}~(b), we employ DINOv2~\cite{dinov2} as our visual backbone $\phi_V$ to extract hierarchical features from both cropped images, \textit{i.e.} $\mathbf{E}^A = \phi_V(\mathbf{I}^A)$, $\mathbf{E}^Q = \phi_V(\mathbf{I}^Q)$. 
The self-supervised pretrained DINOv2 provides robust, generalizable representations that are particularly effective for unseen object categories. For textual understanding, we utilize the CLIP text encoder $\phi_T$~\cite{clip} to convert the text prompt $T$ into a dense embedding $\mathbf{e}^T$.

The extracted visual features $\mathbf{E}^A$ and $\mathbf{E}^Q $ are fused with the text embedding $\mathbf{e}^T$ through a feature fusion module $\phi_{TV}$. $\phi_{TV}$ adopts the same design as the Oryon~\cite{oryon} model. It integrates visual and textual information through a multi-stage process designed to establish robust cross-modal correspondences. This fusion strategy effectively aligns visual features with textual semantics, creating a unified representation.

\begin{figure}[!t]
\centering
\includegraphics[width=0.81\linewidth]{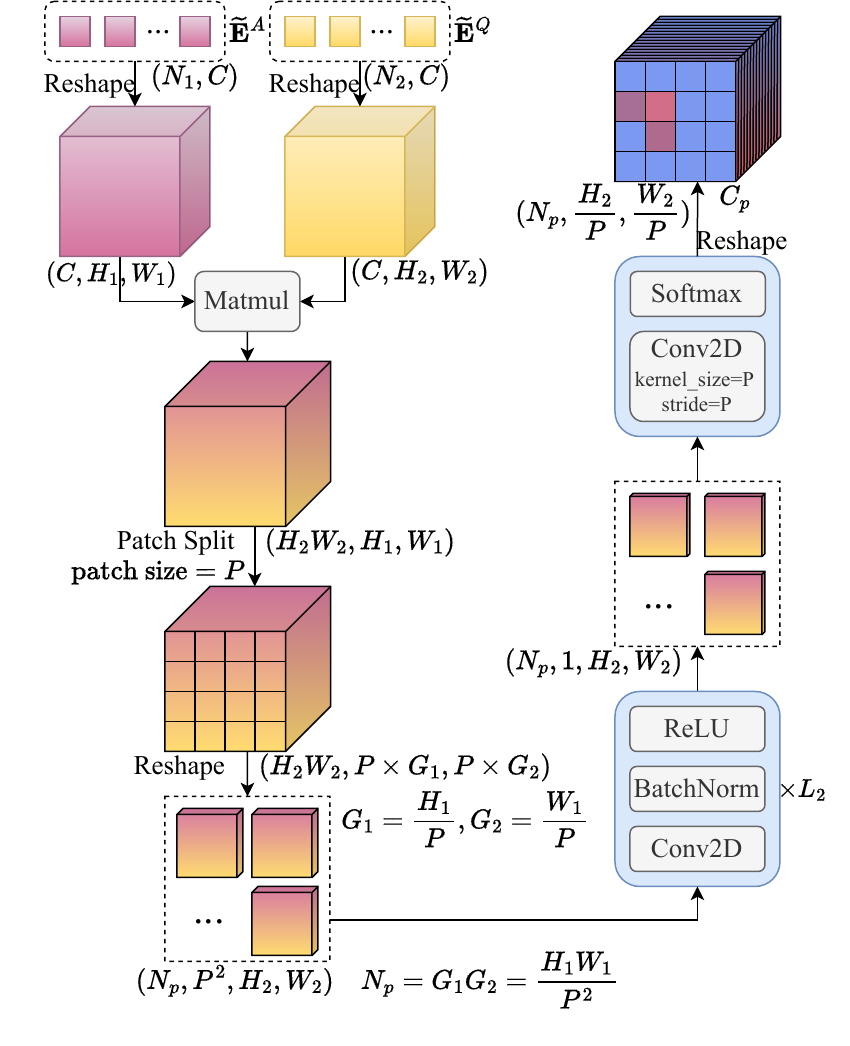}
\vskip-2ex
\caption{Structure of the Patch Correlation Predictor (PCP). 
A patch correlation map is generated through a carefully designed process to establish fine-grained spatial correspondences between anchor and query.}
\label{fig_pcp}
\vskip-2ex
\end{figure}

\subsection{Cross-Perspective Global Perception Module}
To handle drastic viewpoint changes where simple similarity metrics fail, we introduce the Cross-Perspective Global Perception (CPGP) module $\phi_P$. $\phi_P$ establishes robust correspondences between anchor and query views through a multi-layer transformer architecture enhanced with textual guidance. As illustrated in Fig.~\ref{fig_cpgp}, the module receives two fused features $\overline{\mathbf{E}}^A$ and $\overline{\mathbf{E}}^Q$ from the previous stage, which are first combined through a concatenation operation before passing through a down projection layer. The down-projection layer reduces feature channel dimensions while preserving key information and reducing computational complexity. 

The CPGP consists of $L_1$ transformer layers featuring self-attention and cross-attention mechanisms. Self-attention aggregates inter-view information to model the severe 3D geometric deformations and perspective gaps between views. Cross-attention injects textual semantics by using text embeddings as keys and values. This serves as a semantic safeguard, forcing the network to filter out background clutter and focus structural alignment solely on the target object. This architecture enables the model to locate semantically and geometrically analogous regions across drastic viewpoint shifts.

The multi-layer structure allows for progressive refinement of cross-view relationships, where each subsequent layer builds upon the previous one to establish increasingly precise correspondences. After processing through all $L_1$ layers, the enhanced features are projected back to the original number of channels through an up projection layer and separated to produce refined representations $\tilde{\mathbf{E}}^A$ and $\tilde{\mathbf{E}}^Q$ for both views. The output is a pair of structurally aligned feature maps that encode implicitly established global correspondences. This module effectively bridges the perspective gap by learning to align visual features across different perspectives while being guided by textual semantics.

\subsection{Patch Correlation Predictor}
The Patch Correlation Predictor module $\phi_C$ is the core innovation of FiCoP, implementing the coarse-to-fine cognitive mechanism. In open-world 6D pose estimation, unconstrained global matching frequently fails due to background clutter and texture ambiguities. The PCP addresses this task-specific bottleneck by serving as a local matching guide. Instead of allowing unconstrained pixel-wise matching, the PCP generates a patch-to-patch correlation matrix as a spatial structural prior.

The PCP module $\phi_C$ establishes fine-grained spatial correspondences between feature maps from different perspectives through a structured patch-based correlation analysis. As illustrated in Fig.~\ref{fig_pcp}, the module processes two token feature sequences: $\widetilde{\mathbf{E}}^A$ and $\widetilde{\mathbf{E}}^Q$. Firstly, the module reshapes both features into spatial feature maps, then computes their correlation through matrix multiplication to obtain a similarity map of dimensionality $\mathbb{R}^{H_2W_2 \times H_1 \times W_1}$. This similarity map is subsequently partitioned into $G1 \times G2$ non-overlapping patches of size $P \times P$, where each patch represents local correlation patterns between the two views. The patch-split operation transforms the representation into $\mathbb{R}^{N_p \times P^2 \times H_2 \times W_2}$, with $N_p = \frac{H_1W_1}{P^2}$ denoting the total number of patches.

$L_2$ convolutional blocks refine these patch features, followed by a final $P \times P$ convolution and Softmax to produce normalized correlation scores $C_p \in \mathbb{R}^{N_p \times \frac{H_2}{P} \times \frac{W_2}{P}}$. This map explicitly identifies query patches structurally correlated with specific anchor patches, acting as a spatial filter to highlight object parts and suppress clutter. By narrowing the search scope, this step ensures subsequent feature interactions are confined to valid topological regions.

\begin{figure}[!t]
\centering
\includegraphics[width=\linewidth]{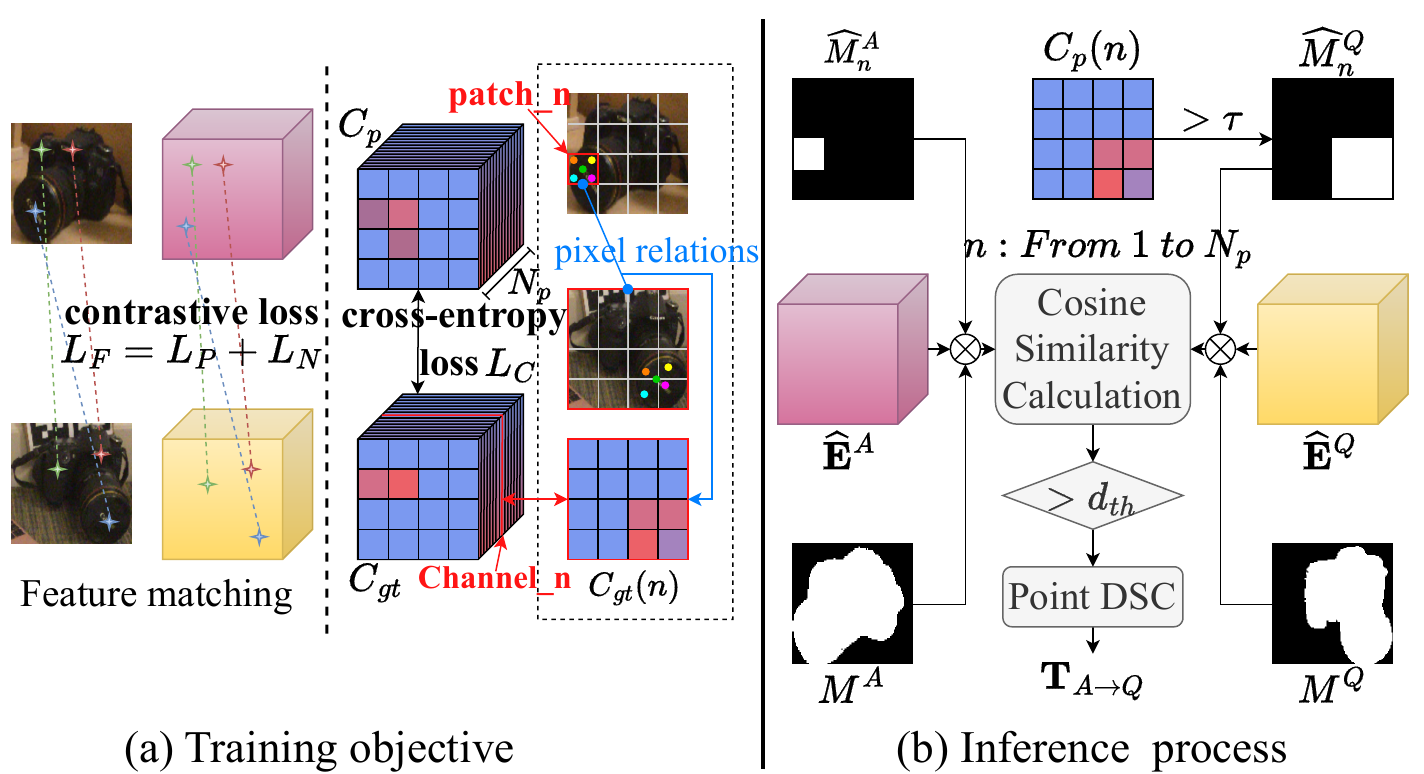}
\vskip-1ex
\caption{Training objectives and inference process. (a) The optimization objective comprises two components: a contrastive loss for feature matching and a classification loss for the patch correlation map. (b) During inference, high-similarity feature pairs are selected from fine-grained regions, and relative poses are computed using the Point DSC algorithm.}
\label{fig_train_test}
\vskip-2ex
\end{figure}

\subsection{Decoder}
To recover pixel-perfect correspondences, the Decoder upsamples the coarse, spatially-constrained features back to the original resolution. The decoder of the FiCoP model shares the same architecture as that of the Oryon model. It consists of three upsampling layers, using $M^A$ or $M^Q$ as guidance, to upsample the fully interactive multimodal features $\widetilde{E}^A$ or $\widetilde{E}^Q$ to the original image resolution, obtaining $\widehat{E}^A$ or $\widehat{E}^Q$ for subsequent dense feature matching.

\subsection{Optimization and Inference Process}
As shown in Fig.~\ref{fig_train_test}~(a), FiCoP incorporates two optimization objectives: feature matching and patch correlation map. For the feature matching loss function, our goal is to maximize similarity between features at matching positions while minimizing similarity at non-matching positions between anchors and queries. We employ the same contrastive loss as Oryon~\cite{oryon}: $\mathcal{L}_{F} = L_P + L_N $. For the patch correlation map, we treat it as a classification problem involving positive and negative samples. The loss is computed using a binary cross-entropy loss function: $\mathcal{L}_{C}  = - \frac{1}{N} \sum_{n=1}^{N} (w_p \cdot C_{gt}(n) \cdot \log(C_p(n)) + (1 - C_{gt}(n)) \cdot  \log(1 - C_p(n)) )$, where $N = N_p \cdot G1 \cdot G2$, $w_p=N_{neg} / {N_{pos}}$ is the positive sample weights. The final total loss function is defined as: $\mathcal{L} = \lambda_1 \mathcal{L}_{F} + \lambda_2 \mathcal{L}_{C}$.

The inference process is shown in Fig.~\ref{fig_train_test}~(b), which uses index $n$ to traverse the number of channels $Np$ in $C_p$, generating the mask $\widehat{M}^A_n$ for the $n$-th patch of the anchor. 
For the query, we binarize $C_p(n)$ to generate the query patch mask $\widehat{M}^Q_n$. 
The binarization method sets patches with values greater than $\tau$ to 1 and the rest to 0. Apply $M^A$ and $\widehat{M}^A_n$ to the feature $\widehat{E}^A$ to filter out the valid anchor feature set $F^A$. Similarly, obtain the valid query feature set $F^Q$. Calculate the cosine similarity between $F^A$ and $F^Q$, adding feature pairs with similarity greater than $d_{th}$ to the matching set. Finally, compute the relative pose $\mathbf{T}_{A \to Q}$ of the matching set using the PointDSC algorithm~\cite{pointdsc}.

\section{Experimental Results}
\subsection{Implementation Details}
Our model was trained on an NVIDIA RTX A6000 GPU with the batch size set to $32$, for a total of $20$ epochs. 
We adopt the Adam optimizer with an initial learning rate of $0.001$ to train our model. 
A cosine annealing scheduler is employed to gradually decay the learning rate throughout training, which helps stabilize convergence and prevent premature overfitting. 

\subsection{Datasets and Metrics}
\subsubsection{Datasets}
Our model is trained on the synthetic ShapeNet6D dataset~\cite{fs6d}, where objects are paired with text descriptions to enable open-vocabulary learning. For evaluation, we strictly adhere to the zero-shot setting, ensuring that objects in the test sets are unseen during training. 
We employ two real-world datasets to assess robustness: REAL275~\cite{nocs} and Toyota-Light~\cite{bop}. 
REAL275 is a benchmark constructed from complex indoor environments. 
Toyota-Light is a dataset specifically designed to evaluate environmental consistency.
Since the open-vocabulary configurations of Linemod and YCB-Video used by Horyon have not yet been released, we strictly evaluate on the publicly accessible REAL275 and Toyota-Light benchmarks to ensure a fair and reproducible comparison.

\begin{table}[!t]
\centering
\renewcommand{\arraystretch}{1.3}
\caption{The comparison results of our method with other methods on the REAL275 and Toyota-Light datasets. The best results are in \textbf{bold}.
\vskip-2ex
\label{table_comparison}}
\begin{tabular}{c|ccc|ccc}
\toprule
\multirow{2}{*}{Method} & \multicolumn{3}{c|}{REAL275}                   & \multicolumn{3}{c}{Toyota-Light}              \\
                        & AR            & ADD           & mIoU          & AR            & ADD           & mIoU          \\ \hline
PoseDiffusion~\cite{posediff}           & 9.5           & 0.8           & -             & 8.1           & 1.6           & -             \\
RelPose++~\cite{realpose++}               & 23.1          & 12.8          & -             & 30.5          & 11.6          & -             \\
ObjectMatch~\cite{objectmatch}             & 21.0          & 11.0          & 81.3          & 8.2           & 4.3           & 82.1          \\
SIFT~\cite{sift}                    & 33.5          & 18.1          & 81.3          & 29.9          & 14.6          & 82.1          \\
LatentFusion~\cite{latentfusion}            & 19.8          & 8.2           & 81.3          & 26.0          & 10.3          & 82.1          \\
Oryon~\cite{oryon}                   & 32.2          & 24.3          & 66.5          & 30.3          & 20.9          & 68.1          \\
Horyon~\cite{horyon}                  & 57.9          & 51.6 & 81.3          & 33.0          & 25.1 & 82.1          \\
Ours                    & \textbf{65.9} & \textbf{55.2}          & \textbf{88.8} & \textbf{39.1} & \textbf{25.6}	& \textbf{89.5} \\ \bottomrule
\end{tabular}
\vskip-3ex
\end{table}

\subsubsection{Evaluation Metrics}
We evaluate predicted poses using Average Recall (AR) and Average Point Distance (ADD). AR integrates three metrics: Visible Surface Discrepancy (VSD), Maximum Symmetry-aware Surface Distance (MSSD), and Maximum Symmetry-aware Projection Distance (MSPD). ADD calculates the average distance between 3D model points under true and estimated poses, using 10\% of the object diameter as the correctness threshold. We also report mean Intersection over Union (mIoU) to evaluate mask quality.

\begin{table*}[thpb]
\centering
\renewcommand{\arraystretch}{1.3}
\caption{The ablation experimental results of our method on the REAL275 and Toyota-Light datasets. $\checkmark$ indicates that this module is being used. The best results are in \textbf{bold}. \label{table_ablation}}
\vskip-2ex
\begin{tabular}{cccc|cccccc|cccccc}
\toprule
\multicolumn{4}{c|}{Model variants} & \multicolumn{6}{c|}{REAL275}                                                                  & \multicolumn{6}{c}{Toyota-Light}                                                              \\
OV-Det       & SAM          & PCP          & CPGP         & AR            & VSD           & MSSD          & MSPD          & ADD           & mIoU      & AR            & VSD           & MSSD          & MSPD          & ADD           & mIoU      \\ \hline
$\checkmark$ &              &              &              & 50.2          & 31.9          & 55.2          & 63.5          & 39.1          & 87.3          & 32.6          & 9.6           & 42.5          & 45.6          & 17.5          & 89.4          \\
$\checkmark$ & $\checkmark$ &              &              & 52.5          & 33.2          & 57.9          & 66.5          & 41.4          & 88.8          & 33.0          & 10.1          & 42.9          & 46.1          & 17.8          & 89.5          \\
$\checkmark$ & $\checkmark$ &              & $\checkmark$ & 62.0          & 43.2          & 69.7          & 73.2          & 46.5          & 88.8          & 36.8          & 12.0          & 47.7          & 50.8          & 20.5          & 89.5          \\
$\checkmark$ & $\checkmark$ & $\checkmark$ &              & 60.9          & 43.1          & 67.4          & 72.1          & 48.7          & 88.8          & 37.1          & 12.3          & 48.4          & 50.7          & 20.5          & 89.5          \\
$\checkmark$ &              & $\checkmark$ & $\checkmark$ & 63.6          & 46.0          & 70.3          & 74.5          & 52.4          & 88.1          & 38.0          & 12.4          & 49.6          & 52.0          & 20.8          & 89.3          \\
             &              & $\checkmark$ & $\checkmark$ & 34.1          & 27.2          & 37.6          & 37.5          & 29.1          & 68.4          & 33.0          & 12.3          & 42.6          & 44.2          & 19.8          & 80.4          \\
$\checkmark$ & $\checkmark$ & $\checkmark$ & $\checkmark$ & \textbf{65.9} & \textbf{48.2} & \textbf{72.7} & \textbf{76.8} & \textbf{55.2} & \textbf{88.8} & \textbf{39.1} & \textbf{13.3} & \textbf{51.0} & \textbf{53.0} & \textbf{25.6} & \textbf{89.5} \\ 
\bottomrule
\end{tabular}
\vskip-3ex
\end{table*}

\subsection{Quantitative Results}
We select seven methods PoseDiffusion~\cite{posediff}, RelPose++~\cite{realpose++}, ObjectMatch~\cite{objectmatch}, SIFT~\cite{sift}, LatentFusion~\cite{latentfusion}, Oryon~\cite{oryon} and Horyon~\cite{horyon} as baseline comparisons. Oryon does not employ cropping preprocessing and utilizes its own predicted object masks. All other methods utilize the open-vocabulary object detection model GroundingDINO for cropping preprocessing. PoseDiffusion and RelPose++ are sparse-view methods that rely solely on RGB data. ObjectMatch, SIFT, and LatentFusion utilize object masks predicted by Horyon.

The data reveals that PoseDiffusion performs poorly on both datasets. ObjectMatch shows slightly better results on REAL275 but performs comparably on Toyota-Light. RelPose++, SIFT, and LatentFusion achieved comparable performance to Oryon on both datasets by employing cropping preprocessing to eliminate most background interference. The novel method Horyon significantly outperforms the aforementioned approaches, demonstrating exceptional competitiveness. Our method achieved the highest score, surpassing all selected baseline methods, thereby proving that our approach has attained state-of-the-art performance in the field of open-vocabulary 6D pose estimation.

\begin{figure}[t]
\centering
\includegraphics[width=0.9\linewidth]{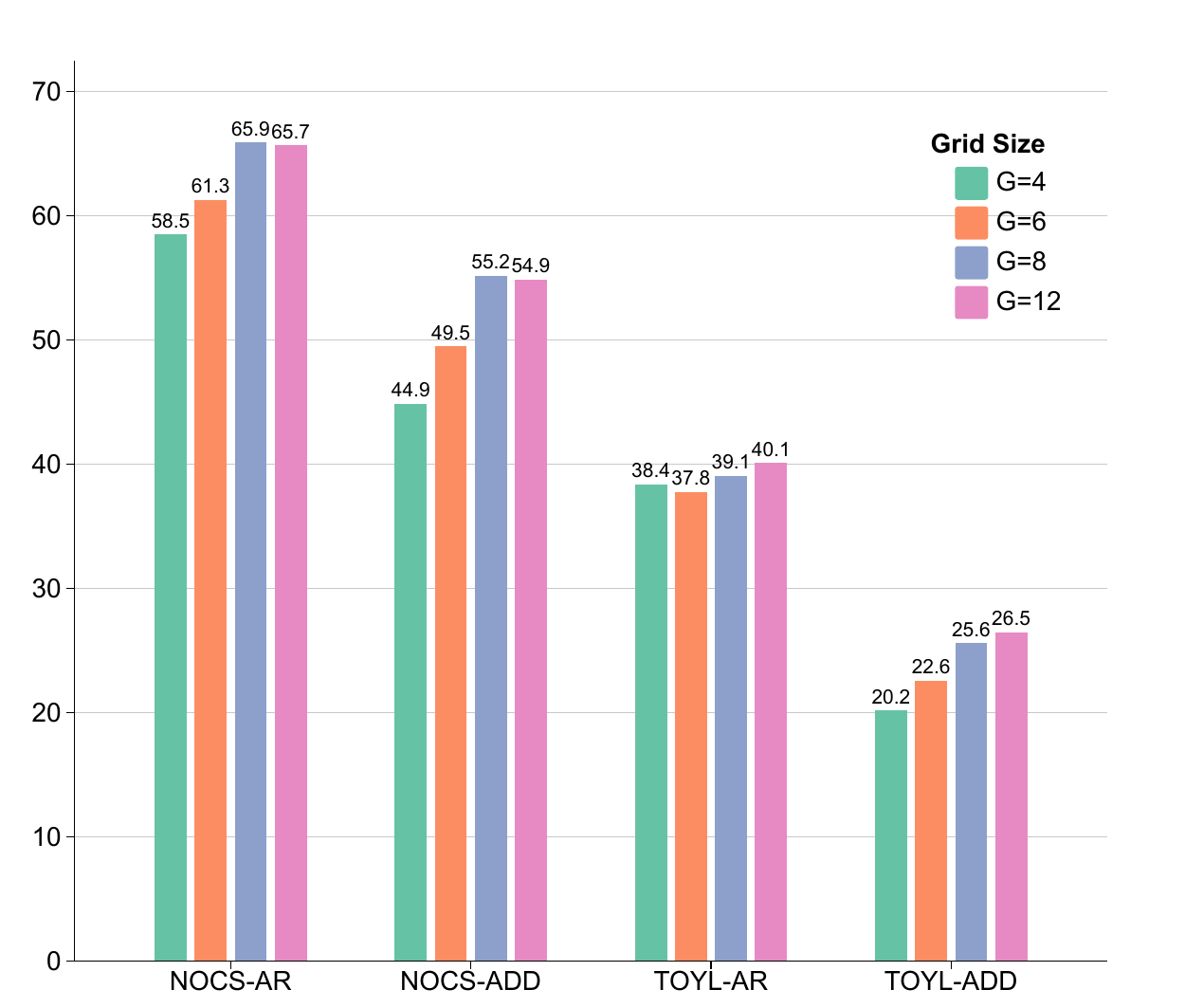}
\vskip-2ex
\caption{Ablation study on the impact of patch granularity on the REAL275 and Toyota-Light datasets. The bar chart reports the Average Recall (AR) and Average Point Distance (ADD) metrics.}
\label{fig_exp_gs}
\end{figure}

\subsection{Ablation Studies}
\subsubsection{Component Analysis}
We evaluate each key component in Table~\ref{table_ablation}. A base model relying on holistic object-level matching achieves 50.2\% AR on REAL275, while adding SAM yields marginal gains to 52.5\%. Individually integrating CPGP or PCP significantly boosts AR to 62.0\% and 60.9\% respectively, demonstrating their strong individual contributions. Their combined integration further surges AR to 65.9\%, confirming our fine-grained correspondence as the primary performance driver. Finally, removing Object-Centric Disentanglement Preprocessing causes the most drastic drop, validating that isolating targets from clutter is a prerequisite for robust pose estimation.

\subsubsection{Impact of Patch Granularity}
We investigate the impact of patch grid size $G \in \{4,6,8,12\}$ as shown in Fig.~\ref{fig_exp_gs}. At $G=4$, performance is lowest as coarse regions fail to filter background clutter effectively. Increasing granularity initially yields significant gains by providing more precise search zones and enforcing structural consistency. However, at $G=12$, performance stabilizes or slightly declines, likely because excessively small patches lack sufficient local semantic context for distinct matching. Balancing stability and efficiency, we adopt $G=8$ as the optimal setting.

\begin{table}[t]
\centering
\renewcommand{\arraystretch}{1.3}
\caption{Optimization results for the binary threshold $\tau$.\label{table_ablation_tau}}
\vskip-2ex
\begin{tabular}{c|cc|cc}
\toprule
\multirow{2}{*}{\hfill Hyperparameter setting \hfill} & \multicolumn{2}{c|}{REAL275} & \multicolumn{2}{c}{Toyota-Light} \\
                        & AR            & ADD           & AR              & ADD            \\ 
\hline
$\tau=0.01$             & 66.1          & 55.1          & 38.2            & 22.1           \\
$\tau=0.02$             & \textbf{66.4} & \textbf{55.6} & 38.6            & 23.8           \\
$\tau=0.03$             & 65.9          & 55.2          & \textbf{39.1}   & 25.1           \\
$\tau=0.04$             & 65.9          & 55.2          & \textbf{39.1}   & \textbf{25.6}  \\
$\tau=0.05$             & 65.7          & 54.9          & 39.0            & 24.8           \\ \bottomrule
\end{tabular}
\vskip-3ex
\end{table}

\begin{table}[t]
\centering
\renewcommand{\arraystretch}{1.3}
\caption{The ablation experimental results of text conditioning.\label{table_ablation_text}}
\vskip-2ex
\begin{tabular}{c|cc|cc}
\toprule
\multirow{2}{*}{Prompt type} & \multicolumn{2}{c|}{REAL275} & \multicolumn{2}{c}{Toyota-Light} \\
                             & AR            & ADD          & AR              & ADD            \\ \hline
Remove prompt fusion         & 62.8          & 51.6         & 37.1            & 22.9           \\
No category prompt           & 63.5          & 52.1         & 38.2            & 23.2           \\
Incorrect attribute prompt   & 65.8          & 54.7         & 38.9            & 24.2           \\
Standard prompt              & \textbf{65.9}           & \textbf{55.2}         & \textbf{39.1}            & \textbf{25.6}           \\ \bottomrule
\end{tabular}
\vskip-2ex
\end{table}

\begin{figure*}[t]
\centering
\includegraphics[width=0.9\linewidth]{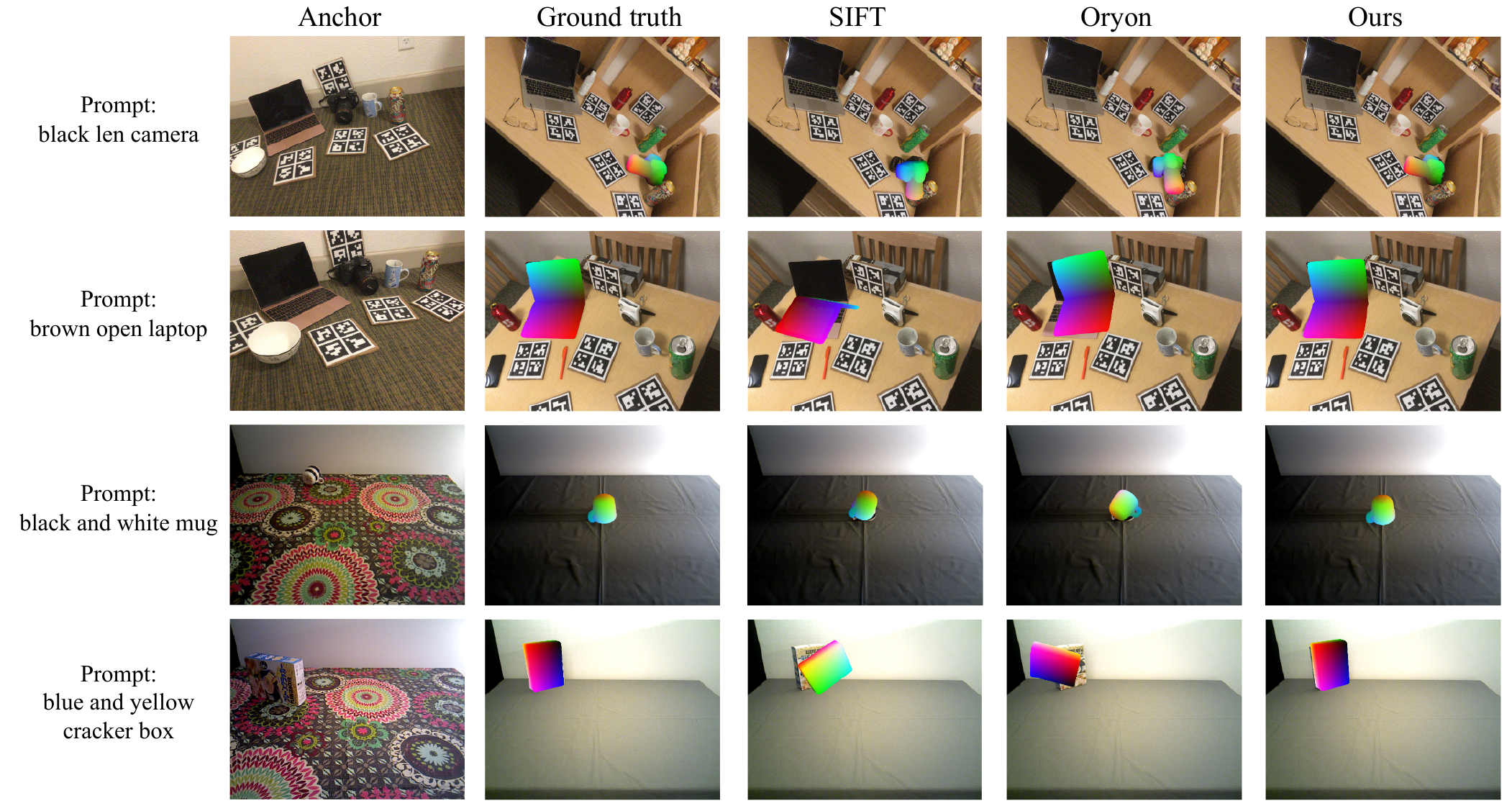}
\vskip-2ex
\caption{Qualitative comparison of predicted poses on the REAL275~\cite{nocs} and Toyota-Light~\cite{bop} benchmarks. The figure shows the poses predicted by SIFT~\cite{sift}, Oryon~\cite{oryon}, and our method. The 3D spatial coordinates of the objects are converted to RGB colors.}
\label{fig_pose_viz}
\end{figure*}

\begin{figure}[t]
\centering
\includegraphics[width=\linewidth]{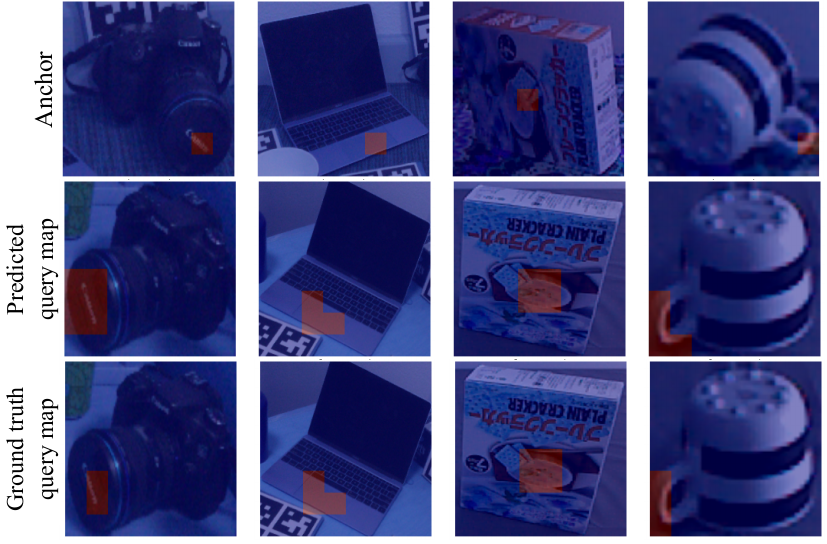}
\vskip-2ex
\caption{Visualization of patch correlation maps. The first column shows a patch from the anchor. The second column displays the relevant patches predicted by the model in the query. The third column presents the ground truth relevant patches.}
\label{fig_patch_viz}
\vskip-2ex
\end{figure}

\begin{figure}[t]
\centering
\includegraphics[width=\linewidth]{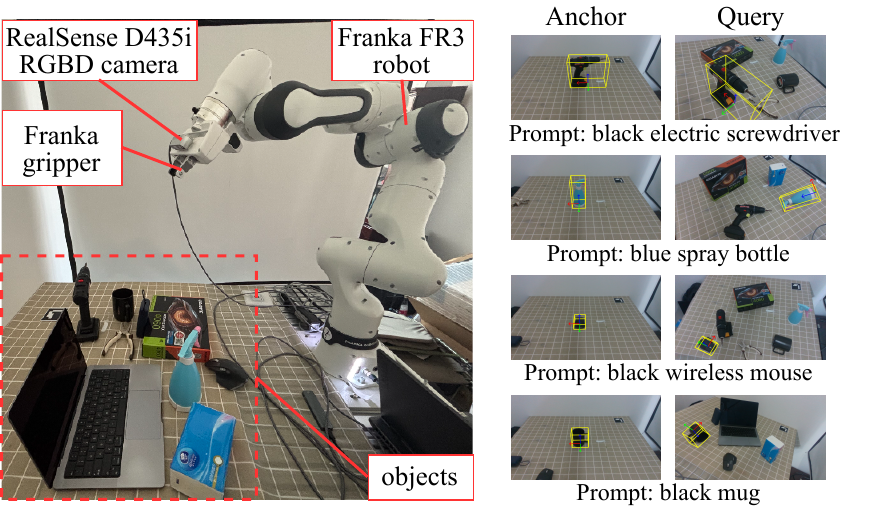}
\vskip-2ex
\caption{Visualization of real-world open-vocabulary 6D pose estimation. Left: The experimental setup featuring a robotic arm equipped with an eye-in-hand RGBD camera in a cluttered tabletop environment. Right: Qualitative results on unseen objects specified by text prompts. The coordinate systems and yellow 3D bounding boxes represent the estimated poses.}
\label{fig_real_world}
\vskip-2ex
\end{figure}

\subsubsection{Sensitivity Analysis of Correlation Threshold}
The binarization threshold $\tau$ acts as a spatial filter to determine valid patch correspondences. As shown in Table~\ref{table_ablation_tau}, varying $\tau$ reveals a trade-off between recall and noise suppression: a low threshold ($\tau=0.01$) admits excessive background clutter, while an aggressive threshold ($\tau=0.05$) risks discarding valid object features. 
Performance remains robust within the range of $[0.02, 0.04]$, peaking at $\tau=0.04$ for Toyota-Light. This confirms that our module effectively learns distinct representations, allowing a simple threshold to reliably separate the object from distractors. We adopt $\tau=0.04$ as the default setting, as it strikes the optimal balance by effectively filtering background noise.

\subsubsection{Effect of Text Conditioning}
We verify the guidance of text features by altering conditions as reported in Table~\ref{table_ablation_text}. The \textbf{Standard prompt} serves as our baseline, utilizing accurate descriptive attributes and specific categories to achieve the optimal performance. In the \textbf{Remove prompt fusion} setting, the text embedding branch is entirely removed from the network, relying solely on visual features. This alteration leads to a significant performance drop, confirming that text acts as an essential semantic anchor for cross-perspective reasoning. Under the \textbf{No category prompt} condition, the specific category name is replaced with the generic word ``object''. This also degrades accuracy, indicating that categorical semantics help disambiguate targets from background distractors. Finally, in the \textbf{Incorrect attribute prompt} setting, the category is correct, but the descriptive attribute is intentionally wrong. Interestingly, this results in only a marginal performance decline. This demonstrates that FiCoP relies predominantly on core structural semantics rather than superficial visual attributes, ensuring stable performance even with imperfect language descriptions.

\begin{table*}[!t]
\centering
\renewcommand{\arraystretch}{1.3}
\caption{Experimental results of the runtime and computational cost of the full pipeline. \label{table_ablation_runtime}}
\vskip-2ex
\begin{tabular}{c|cccc|cccc}
\toprule
\multirow{2}{*}{\begin{tabular}[c]{@{}c@{}}Runtime\\ efficiency\end{tabular}} & \multicolumn{4}{c|}{Initialization process (anchor)}           & \multicolumn{4}{c}{Pose estimation process (query)}                \\
                                     & Pre-processing & Text encoder & Image encoder & Total & Pre-processing & Model inference & Post-processing & Total \\ \hline
Time (ms)          & 702.6                  & 123.7        & 10.2          & 836.5  & 708.6                 & 30.5                  & 257.5           & 996.6  \\
FLOPs (G)          & 600.8                  & 348.7        & 174.7         & 1124.2 & 600.8                 & 589.8                 & -               & 1190.6 \\
Params (M)     & 208.1                  & 56.7         & 302.9         & 567.7  & 208.1                 & 366.9                 & -               & 575.0   \\ \bottomrule
\end{tabular}
\vskip-2ex
\end{table*}

\subsection{Qualitative Results}
\subsubsection{Visualization of Predicted Pose Results}
For a fair comparison, two open-source methods SIFT~\cite{sift} and Oryon~\cite{oryon} are evaluated. 
We run them alongside our method on the same hardware platform to estimate object poses in open-world scenarios using text prompts. 
Visualizing the predicted pose results from all three methods yields the outcome shown in Fig.~\ref{fig_pose_viz}. 
It can be observed that the objects in the four scenes exhibit significant perspective differences between the anchor and query images, presenting considerable difficulty. Consequently, SIFT and Oryon struggle to predict accurate object poses in these scenarios. 
In contrast, our method accurately predicts object poses even in these challenging scenes thanks to the carefully designed fine-grained correspondence strategy.

\subsubsection{Visualization of Patch Correlation Maps}
To explore the role played by patch correlation maps, we visualized the masks generated by their binarization in Fig.~\ref{fig_patch_viz}, overlaid as heatmaps on the RGB image. It can be observed that in the four demonstrated scenarios, despite significant perspective differences between anchors and queries, the patch correlation maps accurately align key regions between anchors and queries, achieving fine-grained correspondence to eliminate interference.

\subsubsection{Visualization in Real-World Scenarios}
We verify the generalization of FiCoP using a robotic manipulator in cluttered real-world environments, as shown in Fig.~\ref{fig_real_world}. Successful 6D pose estimation of diverse, unseen items confirms that FiCoP possesses genuine open-vocabulary capabilities without specific fine-tuning. Results highlight exceptional robustness to large perspective variations, such as frontal to top-down views in the screwdriver and spray bottle cases. Furthermore, the fine-grained correspondence mechanism effectively handles semantic ambiguity by isolating target objects from attribute-similar distractors. This robustness proves the suitability of FiCoP for practical deployment in complex, unconstrained scenarios.

We evaluate the efficiency of FiCoP on an NVIDIA RTX A6000 GPU as detailed in Table~\ref{table_ablation_runtime}. The offline initialization takes 836.5 ms, while online pose estimation requires 996.6 ms per query. The total speed is approximately 1 FPS, which is highly practical for standard robotic manipulation tasks. Since our method provides highly accurate pose estimation, it typically only needs to be computed once prior to grasping, rather than requiring continuous per-frame tracking. The primary bottlenecks of the pipeline are pre-processing foundation models (708.6 ms) and post-processing registration (257.5 ms). Remarkably, the core model inference, including PCP and CPGP modules, requires only 30.5 ms. This demonstrates that our fine-grained correspondence mechanism is computationally lightweight and offers significant potential for real-time acceleration using faster pre-processing foundational models.

\section{Conclusion and Future Work}
We propose FiCoP, a framework for open-vocabulary 6D pose estimation that transitions from noise-prone global matching to spatially-constrained fine-grained correspondence learning. By integrating a Patch Correlation Predictor (PCP) and Cross-Perspective Perception (CPGP), FiCoP leverages structural priors to filter environmental distractors, effectively bridging the gap between VLM semantics and geometric alignment. Experiments on REAL275 and Toyota-Light demonstrate that FiCoP significantly outperforms state-of-the-art baselines under large viewpoint variations and background clutter. Future work will investigate 3D-aware vision-language foundation models and active perception frameworks to autonomously resolve ambiguity in highly occluded environments.

\bibliographystyle{IEEEtran}
\bibliography{reference}

\end{document}